\pdfoutput=1

\documentclass[11pt]{article}

\usepackage[]{EMNLP2022}

\usepackage{times}
\usepackage{latexsym}

\usepackage[T1]{fontenc}

\usepackage[utf8]{inputenc}

\usepackage{microtype}

\usepackage{inconsolata}

\usepackage{algorithm}
\usepackage{algpseudocode}
\usepackage{amsfonts}
\usepackage{amssymb}
\usepackage{fixfoot}
\DeclareFixedFootnote{\repnote}{\url{https://www.nltk.org/_modules/nltk/tokenize/punkt.html}}

\usepackage{graphicx}
\usepackage{multirow}
\usepackage{caption}
\usepackage{booktabs}
\usepackage{amsmath}
\usepackage{comment}
\usepackage{xcolor}
\usepackage[shortlabels]{enumitem}
\usepackage{mathtools}
\usepackage{varioref}
\usepackage{makecell}

\title{Keyphrase Generation Beyond the Boundaries of Title and Abstract}

\author{Krishna Garg \mbox{     }\mbox{     }\mbox{     }\mbox{     } Jishnu Ray Chowdhury \mbox{     }\mbox{     }\mbox{     }\mbox{     } Cornelia Caragea \\
        Computer Science \\ University of Illinois Chicago \\
        \texttt{\{kgarg8,jraych2,cornelia\}@uic.edu}
}

\begin{document}
\maketitle
\begin{abstract}
Keyphrase generation aims at generating important phrases (keyphrases) that best describe a given document. In scholarly domains, current approaches have largely used only the title and abstract of the articles to generate keyphrases. In this paper, we comprehensively explore whether the integration of additional information from the full text of a given article or from semantically similar articles can be helpful for a neural keyphrase generation model or not. We discover that adding sentences from the full text, particularly in the form of the extractive summary of the article can significantly improve the generation of both types of keyphrases that are either present or absent from the text. Experimental results with three widely used models for keyphrase generation along with one of the latest transformer models suitable for longer documents, Longformer Encoder-Decoder (LED) validate the observation. We also present a new large-scale scholarly dataset \textsc{FullTextKP} for keyphrase generation. Unlike prior large-scale datasets, \textsc{FullTextKP} includes the full text of the articles along with the title and abstract. We release the source code at \url{https://github.com/kgarg8/FullTextKP}.
\end{abstract}

\section{Introduction}

Keyphrases of scientific papers provide important topical information about the papers in a highly concise form and are crucial for understanding the evolution of ideas in a scientific field \citep{hall-etal-2008-studying, augenstein-etal-2017-semeval}. Keyphrases of scientific papers can be also useful for mining and analyzing the literature \citep{augenstein-etal-2017-semeval} and for multiple downstream tasks such as index construction \citep{Ritchie:2006:FBI:1629808.1629813}, summarization \citep{Qazvinian:2010:CST:1873781.1873882}, query formulation \citep{song06jcdl}, recommendation \citep{augenstein-etal-2017-semeval}, reviewer matching for paper submissions \citep{augenstein-etal-2017-semeval} and for clustering papers for fast retrieval \citep{Hammouda:2005:CKE:2137868.2137900}. 

Identifying keyphrases from the scientific documents has been popularly studied under the paradigm of Keyphrase Extraction for over a decade \cite{mihalcea2004textrank, Wan:2008:SDK:1620163.1620205, florescu-caragea-2017-positionrank, liu-etal-2009-clustering, Liu:2010:AKE:1870658.1870694, singletpr,sterckx-etal-2016-supervised,caragea-etal-2014-citation,Das_Gollapalli_Caragea_2014,alzaidy2019,patel-caragea-2021-exploiting}. However, one of the major limitations is that these approaches are not able to capture the semantic information in the document, particularly the \textit{absent keyphrases} (keyphrases that are not present in the document).


In this paper, we therefore focus on a newer paradigm of keyphrase generation introduced by \citet{P17-1054}. Instead of extracting keyphrases from the document, we model the problem as a neural sequence-to-sequence (seq2seq) approach where we learn to generate keyphrases in an autoregressive way.
This approach can not only predict the exact keyphrases that are present in a document ({\em present keyphrases}) but also those that are semantically relevant but absent from the document ({\em absent keyphrases}). While many works \citep{P17-1054, chen2019guided, chen-etal-2020-exclusive, yuan-etal-2020-one, ye2021one2set} adopt this approach, their focus is on architectural innovations to improve the generation of both present and absent keyphrases. In contrast, our focus is different from that of the prior works in keyphrase generation: we propose to extend the source of information of the input sequence, which is thus far limited to only title and abstract.

\begin{table*}[t]
\centering
\def\columnseprulecolor{\color{grey}}
\resizebox{\textwidth}{!}{
  \begin{tabular}{l|p{47.5em}}
    \toprule
   \multicolumn{1}{c}{\rule{0pt}{2ex}\textbf{Method}} &   \multicolumn{1}{c}{\textbf{Snippets from title and abstracts and beyond}}\\[1ex] 
    \hline
    \rule{0pt}{3ex}Title & Performance of \colorbox{cyan!25!white}{Power Detector} Sensors of DTV Signals in \colorbox{cyan!25!white}{IEEE 802.22} WRANs\\[0.5ex]
    \hline
     \rule{0pt}{3ex}Abstract &  Sensing is the most important component in any \colorbox{cyan!25!white}{cognitive radio} system. The \colorbox{cyan!25!white}{IEEE 802.22} Working Group (WG) is formulating the first worldwide standard for \colorbox{cyan!25!white}{cognitive radios} to operate in the television (TV) bands...\\[0.5ex]
    \hline
    \rule{0pt}{3ex}Citations & The \colorbox{cyan!25!white}{IEEE 802.22} functional requirements document [2] states that \colorbox{cyan!25!white}{spectrum sensing} is required ...\\[0.5ex]
    \hline
    \rule{0pt}{3ex}Non-Citations & The \colorbox{cyan!25!white}{keep-out region} is a region around the primary user (e.g. DTV transmitter) where... Given these simulation scenarios we can evaluate various \colorbox{cyan!25!white}{spectrum sensing} techniques.\\
    \hline
    \rule{0pt}{3ex}Summary & Of course, identification of which TV channels... This paper will describe ... in the \colorbox{cyan!25!white}{IEEE 802.22} WG to evaluate \colorbox{cyan!25!white}{spectrum sensing} techniques... Calculating the base station \colorbox{cyan!25!white}{keep-out region} ...\\
    \hline
    \rule{0pt}{3ex}Ret-Aug & The Bernoulli nonuniform sampling is further extended to matrix formulation, which allows the application of \colorbox{cyan!25!white}{spectrum sensing} for \colorbox{cyan!25!white}{cognitive radio} signal detection.
    
    \textit{- From Paper: 'Noise Enhancement and SNR Equivalence in Bernoulli Nonuniform Sampling'}\\
    \hline
    \rule{0pt}{3ex}Keyphrases & \colorbox{cyan!25!white}{Cognitive Radio}, \colorbox{cyan!25!white}{IEEE 802.22}, \colorbox{cyan!25!white}{Spectrum Sensing}, \colorbox{cyan!25!white}{Power Detector}, \colorbox{cyan!25!white}{Keep-out Region}, \colorbox{red!25!white}{False alarm probability}, \colorbox{red!25!white}{Misdetection probability}\\
    \bottomrule
  \end{tabular}
}
  \caption{A sample from \textsc{FulltextKP} dataset showing \colorbox{cyan!25!white}{\textsc{Present}} and \colorbox{red!25!white}{\textsc{Absent}} gold keyphrases in the different semantic spaces of the article. Refer to the Methods section for the description of the methods.}
    \label{tab:cherryexample}
    
\end{table*}

To this end, we curate a large-scale dataset of papers published by ACM\footnote{\url{https://dl.acm.org/}} which contains not only the title and abstract of the documents but also the full text from the documents. We call this dataset \textsc{FullTextKP}. Owing to the scarcity of such large-scale datasets with full text and the difficulty of parsing and understanding the entire text of the documents, the integration of additional information that goes beyond the title and abstract for keyphrase generation has been largely ignored. 

In this paper, we provide innovative ways of using certain parts of the documents that could be rich source of information, e.g., using citation sentences from the content of the document or an extractive summary of the document (as illustrated in Table \ref{tab:cherryexample}). Interestingly, through comprehensive experiments, we find that the citation contexts or provenance that had been a rich source of information for the keyphrase extraction task
\cite{caragea-etal-2014-citation, Das_Gollapalli_Caragea_2014} are not the richest sources of information in the keyphrase generation task. In contrast, we observe that the semantic information from the document could be best assimilated if we summarize the document and use the summary of the document instead of the document itself. To summarize each document, we use an unsupervised summarization approach and show that the extractive summmary not only contains the highest topical information useful for keyphrase generation but can also fit the computational budget. Remarkably, the approach based on the incorporation of summary outperforms methods that use just title and abstract by a wide margin, oftentimes yielding 2-3 times improvements in performance in both present and absent keyphrases. 

Our experiments are performed with three widely used models for keyphrase generation as well as, to our knowledge, for the first time in keyphrase generation, with a transformer model suited for long documents, Longformer  Encoder-Decoder (LED) \cite{Beltagy2020Longformer}.

Overall, we make the following contributions:
\vspace{-1mm}
\begin{enumerate}
    \item We 
    explore the benefit of integrating additional information from different data sources (not just the title and abstract) into neural seq2seq models for keyphrase generation (i.e., predicting both present and absent keyphrases). 
    The different data sources include random sentences from the article body,\footnote{\textit{Body} includes all the sections of a scientific article besides the title and abstract} sentences from the summary of the body, citation sentences, non-citation sentences, and sentences from other related documents in the training set. 
    \vspace{-1mm}
    \item We show that the sentences from the body of the article that form an extractive summary are great sources of finding good content for keyphrase generation.
    \vspace{-1mm}
    \item We present a new dataset, \textsc{FullTextKP}, of $\sim$140,000 articles, each with its full text.
\end{enumerate}

\section{Related Work}

\textbf{Keyphrase Generation.} Recent approaches to keyphrase generation have been dominated by neural seq2seq models because they provide a mechanism to also generate absent keyphrases. \citeauthor{P17-1054} \shortcite{P17-1054} originally proposed an RNN-based seq2seq model along with CopyNet (CopyRNN) for keyphrase generation. \citeauthor{chen-etal-2018-keyphrase} \shortcite{chen-etal-2018-keyphrase} improved CopyRNN by taking correlations between the predicted keyphrases into account (CorrRNN). 
\citet{yuan-etal-2020-one} extended the CopyRNN model to dynamically generate a variable number of diverse keyphrases.
\citet{chen-etal-2020-exclusive} proposed an exclusive hierarchical decoding framework to capture the hierarchical compositionality of keyphrases (ExHiRD). \citet{swaminathan-etal-2020-preliminary} explored Generative Adversarial Networks for keyphrase generation. \citet{ye2021one2set} modeled the target keyphrases as a set instead of a predefined sequence to improve the biased matching of the generated and target keyphrases. These methods use only title and abstract as the input sequence. Our work can aid all of these methods to further enhance the performance of keyphrase generation by augmenting the source texts with information from the body of the documents.

\textbf{External Information.} Some of the previous methods try to incorporate external information to enhance the performance of keyphrase extraction. For example, MAUI \cite{Medelyan:2009:HTU:1699648.1699678} uses semantic information from Wikipedia,  CeKE \cite{caragea-etal-2014-citation} and CiteTextRank \cite{GollapalliC14} use information from  citation networks, and SingleRank and ExpandRank \cite{Wan:2008:SDK:1620163.1620205} use information from local textual neighborhoods. However, these methods are limited to extraction-based methods whereas we focus on generating keyphrases to have a holistic or semantic understanding of the document when the natural language text is fed to the neural model.

Several works in keyphrase generation started to use external information to generate keyphrases.  
For example, \citet{chen-etal-2019-integrated}, 
\citet{santosh2021gazetteer} proposed methods where they augmented a set of additional keyphrases from semantically similar documents and use hidden state representations built from these keyphrases, \citet{diao2020keyphrase} used cross-document attention networks, \citet{ye2021heterogeneous} used information from retrieved document-keyphrase pairs that are similar to the source document,
\citet{shen2021unsupervised} used a phrase bank by pooling all keyphrases from all the articles. Similar to these works, one of our methods (Retrieval-Augmentation) 
retrieves relevant sentences from semantically similar documents and directly appends them to the title and abstract. 

\textbf{Datasets.} There have been numerous efforts in the direction of curating large-scale datasets for either keyphrase extraction or keyphrase generation. In the non-scholarly domain, the datasets such as KPTimes \citep{gallina-etal-2019-kptimes}, JPTimes \citep{gallina-etal-2019-kptimes}, KPCrowd \citep{marujo2013supervised}, DUC-2001 \citep{wan2008single}, were curated from news articles, OpenKP \citep{xiong-etal-2019-open} was mined from webpages from a search engine. In the scholarly domain, there has been ongoing research with the datasets such as Krapivin \citep{key:dataset2009krapivin-autayeu-marchese}, Inspec \citep{hulth2003improved}, SemEval \citep{Kim:2010:STA:1859664.1859668}, NUS \citep{nguyen2007keyphrase}, KP20k \citep{P17-1054} and OAGK \citep{cano2019keyphrase}. The limitation of these datasets is that they are either very small (of the magnitude of a few hundreds or thousands) or they are bounded to title and abstract. For instance, the widely used KP20k dataset \citep{P17-1054} consists of only the title and abstract of the research articles and the average words per document is as low as 176. To leverage and understand the benefit of additional data, we mined a new dataset based on the articles primarily from the ACM database which contains not just the titles and abstracts but also the full text of the scholarly documents. In our dataset, the typical length of full papers is 5000-10000 tokens.

\section{Methods}

\label{sec:methods}

We use four different models, three of which are popular models in the Keyphrase Generation task, viz. catSeq \cite{yuan-etal-2020-one}, One2Set \cite{ye2021one2set}, ExHiRD \cite{chen-etal-2020-exclusive}; and the fourth model is based on one of the latest Transformer models, i.e., Longformer Encoder-Decoder (LED) \cite{Beltagy2020Longformer}, suitable for long documents. The baseline in each model takes the title and abstract (T+A) as input and predicts a sequence of keyphrases. We explore different extensions (also illustrated in Table \ref{tab:cherryexample}) for each model by concatenating different types of data to the title and abstract (\textbf{T+A}). During the concatenation of additional information, we use a delimiter $<$sep$>$ between the title and abstract and between all the additional sentences as follows: Title $<$sep$>$ Abstract $<$sep$>$ Sent$_1$ $<$sep$>$ Sent$_2$  $<$sep$>$ ... Sent$_k$. We describe different types of data (input sequence) below:

\vspace{1mm}
\noindent \textbf{\textsc{T+A+Random:}} For this setup, we concatenate $k$ randomly chosen sentences (from the body) to the title and abstract in the original order of their occurrence in the body.

\vspace{1mm}
\noindent \textbf{\textsc{T+A+Citations:}} For this setup, we collect all the sentences (from the body) that have cited some other article. We call such sentences as \textit{citation sentences}. We randomly select $k$ sentences from the entire pool and concatenate them to the title and abstract in the order of their occurrence in the body of the given article. The incorporation of citation sentences is inspired by Na\"ive-Bayes or TextRank-based approaches \cite{GollapalliC14, caragea-etal-2014-citation} or construction of CitationGraph \cite{CitationIE} which enhanced their keyphrase extraction performance through the integration of citation information.
\vspace{1mm}

\noindent \textbf{\textsc{T+A+Non-Citations:}} For this setup, we collect all the sentences (from the body) that do not cite any other article. We randomly select $k$ sentences from the entire pool and concatenate them to the title and abstract in the order of their occurrence in the body of the given article. 

\vspace{1mm}
\noindent \textbf{\textsc{T+A+Summary:}} In this method, we summarize the body of the article using a state-of-the-art unsupervised summarization algorithm, \textit{PacSum} \citep{zheng2019sentence} and append $k$ sentences from the summary to T+A of the article. 
\citeauthor{zheng2019sentence} showed that TF-IDF-based PacSum is better than TextRank and other baselines, and is also highly competitive with even BERT-based PacSum. PacSum can be replaced with other summarization algorithms but we choose TF-IDF-based PacSum simply as a representative of a powerful and efficient model for unsupervised summarization to test the efficacy of summarizing the body.

The steps to obtain the extractive summary of the documents using PacSum are as follows. Consider a document as a list of sentences and each sentence as a list of words. The entire document can be visualized as a directed graph in which the individual sentences are the nodes and the edges are weighted by similarity. 
We first calculate the tf-idf to understand the relevance of a word to a sentence in the collection of sentences (document), quite analogous to computing tf-idf of a word of a document in a corpus of documents. We then compute the similarity between every pair of sentences. Next, we compute a threshold based on which we normalize the similarity scores. Subsequently, we compute forward and backward edge scores for the directed edges. Afterward, we compute the degree centrality of a sentence (node) in the document (graph) as the weighted average of forward and backward edge scores. Finally, we rank the sentences based on their centrality values, select the $k$ highest ranked sentences, and concatenate them to the title and abstract. We describe the algorithm in Appendix \ref{sec:PacSum}.

\vspace{1mm}

\noindent \textbf{\textsc{T+A+Retrieval-Augmentation:}} In this method, we retrieve and augment $k$ semantically similar sentences from the training corpus to the T+A of each article. To this end, we first create a set of all the sentences from the titles and abstracts of the articles in the entire \textit{training} dataset. Next, we embed each sentence in the set using \textsc{SPECTER} \citep{cohan2020specter}. We treat these embeddings as \textit{key embeddings} representing the corresponding sentences (\textit{values}). Given a target article (\textit{query}), we then embed its title and abstract using \textsc{SPECTER}. This embedding serves as a \textit{query embedding} to search related sentences from other articles. Subsequently, we compute the dot-product \textit{similarity} of the query embedding with all the key embeddings using \textsc{FAISS} \citep{johnson2019billion}. 
Last, we select $k$ sentences (values) corresponding to the top $k$ most similar key embeddings and concatenate the sentences to the title and abstract of the query article. 
We provide the algorithm in Appendix \ref{sec:ret-aug}.

\section{Experimental Setup}

\subsection{ \textsc{FullTextKP} dataset}

\noindent 
To evaluate the performance of models leveraging different types of information, we construct a new dataset, which we call \textit{\textsc{FullTextKP}}.
Our \textsc{FullTextKP} dataset consists of research papers that are published by ACM\footnote{The full ACM dataset is available from the ACM Digital Library by request.} and are available in the ACM digital library 
in its International Conference Proceedings Series (ICPS). 

We used only the articles which have at least the five fields, viz., title, abstract, keywords, full text and references. We lowercased the text and constructed numerous regexes to remove the escape sequences, html tags, urls, emails, etc. We also replaced all the numbers and roman numerals with a <digit> token. 
Further, we used PunktSentenceTokenizer\repnote\ to segment the document into sentences and NLTK's word\_tokenizer\repnote\ to tokenize the sentences into tokens. We constructed suitable regex to extract the citation sentences and we considered all other sentences as non-citation sentences.
The articles without any citation sentences and duplicates were further removed from the collection. Thus, we were able to collect 142,844 articles. We split the \textsc{FullTextKP} dataset into 80/10/10 for train, test and validation sets. Statistical information about the dataset is shown in Table \ref{tab:statistics}.

The construction of this dataset addresses the sparsity of large-scale datasets for keyphrase extraction/generation from scientific papers and is aimed at enabling deep learning modeling. Currently, there exist only a couple of large-scale datasets for this task, e.g., \textsc{KP20k} \cite{P17-1054}, and \textsc{OAGK} \cite{cano2019keyphrase}. Unlike these datasets, which contain only the title and abstract of each paper, our dataset provides access to the full text of each paper. 

\begin{table}[!]
\small
\centering
\def\arraystretch{1.2}
\begin{tabular}{  p{17em} | r } 
\Xhline{2\arrayrulewidth}
\hline
\addlinespace[1ex]
Total papers in the corpus & $142,844$\\
Training set size & $114,271$\\
Validation set size &  $14,287$\\
Test set size & $14,286$\\
\% of present keyphrases & $55.8$\\
\% of absent keyphrases & $44.2$\\
Average keyphrases per paper & $4.3$\\
\addlinespace[1ex]
\Xhline{2\arrayrulewidth}
\hline
\end{tabular}
\caption{\textsc{FulltextKP} dataset statistics}
\label{tab:statistics}
\end{table}

\begin{table*}[ht]
\small
\centering
\renewcommand{\arraystretch}{1.1}
\setlength\tabcolsep{13pt}
\begin{tabular}{clcccccc} 
\Xhline{3\arrayrulewidth}
\hline
Model & Method & P@5 & R@5 & F1@5 & P@M & R@M & F1@M \\ 
\hline
\multirow{6}{*}{catSeq} & Title+Abstract & 0.167$_2$ & 0.343$_3$ & 0.224$_2$ & 0.333$_3$ & 0.343$_4$ & 0.338$_1$ \\
& + Ret-Aug & 0.166$_0$ & 0.342$_1$ & 0.224$_0$~ & 0.338$_3$ & 0.343$_1$ & 0.340$_1$ \\
& + Citations & 0.170$_1$ & 0.351$_3$~ & 0.229$_2$~ & 0.360$_2$ & 0.351$_3$ & 0.355$_1$ \\
& + Non-Citations & 0.178$_1$ & 0.365$_3$~ & 0.240$_2$ & 0.365$_3$ & 0.366$_3$ & 0.366$_2$ \\
& + Random & 0.175$_2$ & 0.359$_5$~ & 0.236$_3$~ & 0.370$_4$ & 0.359$_5$ & 0.364$_4$ \\
& + Summary & \textbf{0.187$_1$} & \textbf{0.380$_2$} & \textbf{0.250$_2$}~ & \textbf{0.379$_4$} & \textbf{0.380$_2$} & \textbf{0.380$_2$} \\
\hline

\multirow{6}{*}{One2Set} & Title+Abstract & \textbf{0.203$_3$} & \textbf{0.418$_4$} & \textbf{0.273$_3$} & 0.300$_6$ & \textbf{0.425$_6$} & 0.351$_2$ \\
& + Ret-Aug & 0.178$_4$ & 0.368$_9$ & 0.241$_5$ & 0.313$_5$ & 0.370$_9$ & 0.339$_3$ \\
& + Citations & 0.192$_1$ & 0.394$_4$ & 0.258$_2$ & 0.316$_5$ & 0.396$_3$ & 0.351$_1$ \\
& + Non-Citations & 0.192$_6$ & 0.395$_{13}$ & 0.259$_9$ & 0.316$_6$ & 0.397$_{14}$ & 0.352$_2$ \\
& + Random & 0.191$_4$ & 0.391$_7$ & 0.257$_4$ & 0.314$_9$ & 0.394$_8$ & 0.349$_3$ \\
& + Summary & 0.189$_5$ & 0.385$_{13}$ & 0.253$_9$ & \textbf{0.325$_7$} & 0.387$_{14}$ & \textbf{0.353$_2$} \\
\hline

\multirow{6}{*}{ExHiRD} & Title+Abstract & 0.184$_2$ & 0.381$_3$ & 0.236$_2$ & 0.320$_3$ & 0.382$_3$ & 0.325$_1$ \\
& + Ret-Aug & 0.187$_0$ & 0.390$_2$ & 0.241$_1$ & 0.317$_2$ & 0.391$_2$ & 0.325$_2$ \\
& + Citations & 0.194$_0$ & 0.402$_1$ & 0.249$_0$ & 0.336$_1$ & 0.403$_1$ & 0.340$_0$ \\
& + Non-Citations & 0.197$_1$ & 0.410$_6$ & 0.257$_8$ & 0.353$_2$ & 0.407$_1$ & 0.354$_1$ \\
& + Random & 0.195$_0$ & 0.403$_1$ & 0.249$_0$ & 0.352$_2$ & 0.404$_1$ & 0.351$_1$ \\
& + Summary & \textbf{0.202$_2$} & \textbf{0.416$_3$} & \textbf{0.258$_2$} & \textbf{0.369$_1$} & \textbf{0.416$_4$} & \textbf{0.366$_0$} \\
\hline

\multirow{6}{*}{LED
} & Title+Abstract & 0.193$_{11}$ & 0.389$_{21}$ & 0.258$_{15}$ & 0.333$_1$ & 0.392$_{22}$ & 0.360$_9$ \\
& + Ret-Aug & 0.193$_6$ & 0.391$_{10}$ & 0.258$_7$ & 0.319$_4$ & 0.394$_{11}$ & 0.353$_2$ \\
& + Citations & 0.201$_5$ & 0.403$_9$ & 0.268$_6$ & 0.338$_4$ & 0.407$_{10}$ & 0.369$_2$ \\
& + Non-Citations & 0.209$_3$ & 0.421$_6$ & 0.280$_4$ & 0.348$_9$ & 0.425$_7$ & 0.382$_3$ \\
& + Random & 0.213$_0$ & 0.428$_1$ & 0.284$_0$ & 0.354$_6$ & 0.431$_0$ & 0.389$_3$ \\
& + Summary & \textbf{0.217$_4$} & \textbf{0.431$_8$} & \textbf{0.289$_5$} & \textbf{0.365$_2$} & \textbf{0.435$_8$} & \textbf{0.397$_3$} \\


\Xhline{3\arrayrulewidth}
\hline
\end{tabular}

\caption{Results comparing present keyphrase performance with different methods of using additional data with the \textsc{FullTextKP} dataset. Subscripts denote the standard deviation as multiples of $\pm$ 0.001.}
\label{tab:main-table}
\end{table*}

\begin{table*}[ht]
\small
\centering
\renewcommand{\arraystretch}{1.1}
\setlength\tabcolsep{13pt}
\begin{tabular}{clcccccc} 
\hline
\Xhline{3\arrayrulewidth}
Model & Method & P@5 & R@5 & F1@5 & P@M & R@M & F1@M \\ 
\hline
\multirow{6}{*}{catSeq} & Title+Abstract & 0.007$_0$ & 0.017$_1$~ & 0.010$_1$~ & 0.030$_2$ & 0.017$_1$ & 0.021$_1$ \\
& + Ret-Aug & 0.009$_0$ & 0.021$_1$~ & 0.012$_1$~ & 0.037$_1$ & 0.021$_1$ & 0.027$_1$ \\
& + Citations & 0.013$_1$ & 0.031$_2$~ & 0.019$_2$~ & 0.052$_3$ & 0.031$_2$ & 0.039$_3$ \\
& + Non-Citations & 0.021$_0$ & 0.046$_1$~ & 0.029$_0$~ & 0.068$_1$ & 0.046$_1$ & 0.055$_0$ \\
& + Random & 0.022$_1$ & 0.049$_2$ & 0.030$_1$~ & 0.073$_3$ & 0.049$_2$ & 0.059$_3$ \\
& + Summary & \textbf{0.033$_1$} & \textbf{0.069$_2$}~ & \textbf{0.045$_1$}~ & \textbf{0.093$_2$} & \textbf{0.069$_2$} & \textbf{0.079$_2$} \\
\hline

\multirow{6}{*}{One2Set} & Title+Abstract & 0.018$_1$ & 0.042$_2$ & 0.026$_1$ & 0.043$_1$ & 0.042$_2$ & 0.043$_0$ \\
& + Ret-Aug & 0.017$_0$ & 0.039$_2$ & 0.023$_1$ & 0.044$_2$ & 0.039$_2$ & 0.041$_1$ \\
& + Citations & 0.021$_1$ & 0.047$_2$ & 0.029$_1$ & 0.053$_1$ & 0.047$_2$ & 0.049$_1$ \\
& + Non-Citations &  0.023$_1$ & \textbf{0.053$_2$} & 0.032$_1$ & 0.054$_1$ & \textbf{0.053$_2$} & 0.053$_2$ \\
& + Random & \textbf{0.024$_0$} & \textbf{0.053$_2$} & \textbf{0.033$_1$} & \textbf{0.056$_1$} & \textbf{0.053$_2$} & \textbf{0.054$_2$} \\
& + Summary & 0.023$_3$ & 0.050$_6$ & 0.032$_4$ & 0.054$_4$ & 0.051$_6$ & 0.052$_5$ \\
\hline

\multirow{6}{*}{ExHiRD} & Title+Abstract & 0.012$_0$ & 0.027$_2$ & 0.016$_1$ & 0.042$_1$ & 0.026$_0$ & 0.030$_0$ \\
& + Ret-Aug & 0.010$_0$ & 0.022$_0$ & 0.013$_0$ & 0.035$_1$ & 0.022$_0$ & 0.025$_0$ \\
& + Citations & 0.019$_0$ & 0.041$_0$ & 0.024$_0$ & 0.060$_1$ & 0.041$_0$ & 0.045$_1$ \\
& + Non-Citations & 0.024$_0$ & 0.051$_0$ & 0.031$_0$ & 0.071$_1$ & 0.051$_0$ & 0.055$_0$ \\
& + Random & 0.026$_0$ & 0.054$_1$ & 0.032$_0$ & 0.076$_1$ & 0.054$_1$ & 0.059$_1$ \\
& + Summary & \textbf{0.037$_1$} & \textbf{0.074$_3$} & \textbf{0.046$_1$} & \textbf{0.100$_2$} & \textbf{0.074$_3$} & \textbf{0.080$_2$} \\
\hline

\multirow{6}{*}{LED
} & Title+Abstract & 0.027$_0$ & 0.063$_1$ & 0.038$_0$ & 0.061$_{11}$ & 0.063$_1$ & 0.061$_5$ \\
& + Ret-Aug & 0.014$_1$ & 0.034$_2$ & 0.020$_1$ & 0.053$_1$ & 0.034$_2$ & 0.041$_1$ \\
& + Citations & 0.025$_2$ & 0.058$_4$ & 0.035$_3$ & 0.073$_2$ & 0.058$_4$ & 0.065$_3$ \\
& + Non-Citations & 0.033$_1$ & 0.071$_2$ & 0.045$_1$ & 0.088$_1$ & 0.071$_2$ & 0.079$_1$ \\
& + Random & 0.036$_2$ & 0.078$_4$ & 0.049$_2$ & 0.093$_1$ & 0.078$_4$ & 0.085$_2$ \\
& + Summary & \textbf{0.047$_1$} & \textbf{0.100$_1$} & \textbf{0.064$_1$} & \textbf{0.112$_4$} & \textbf{0.100$_1$} & \textbf{0.106$_2$} \\


\Xhline{3\arrayrulewidth}
\end{tabular}

\caption{Results comparing absent keyphrase performance with different methods of using additional data with the \textsc{FullTextKP} dataset. Subscripts denote the standard deviation as multiples of $\pm$ 0.001.}
\label{tab:main-table2}
\end{table*}

\subsection{Evaluation}
\label{Evaluation}

We use the same evaluation method as \citet{chan-etal-2019-neural,chen-etal-2020-exclusive} and report the macro-averaged Precision ({\bf P}), Recall ({\bf R}) and F1-scores ($\bf F_1$) in two different evaluation settings: \textbf{@5} (dummy keyphrases are appended to the predicted keyphrases to make the total count as 5) and \textbf{@M} (directly compare the generated keyphrases against the gold keyphrases). All the keyphrases are stemmed using PorterStemmer\footnote{\url{https://www.nltk.org/\_modules/nltk/stem/porter.html}} before comparison. As in prior works, we treat a keyphrase as \textit{absent} if it is absent from the title and abstract, 
which helps us compare the performance of generating \textit{absent} and \textit{present} keyphrases for different models in a consistent way.

For the entire data including Title, Abstract and the additional data using different methods, we restrict the maximum sequence length to $800$. This requires us to use an adaptive $k$ for the additional number of sentences that can fit in the given maximum sequence length.
The selection of the sequence length was based on the tradeoff between the performance and computational budget of the models. We also experiment with longer sequence lengths for the LED model in \S\ref{sec:LED-exploration}.

\section{Results}
In this section, we first present the results for present and absent keyphrase generation performance in \S \ref{sec:keyphrase-performance} for the different methods described in \S \ref{sec:methods}. We further explore our best performing method, i.e., T+A+summary into two sub-categories: extractive and abstractive summarization methods in \S \ref{sec:summarization}. Finally, in \S \ref{sec:LED-exploration} we explore the performance of the best performing model, i.e., Longformer Encoder-Decoder with sequence lengths much higher than 800. We run all the experiments three times and report the average.

\subsection{Present \& Absent Keyphrase Generation}
\label{sec:keyphrase-performance}


Tables \ref{tab:main-table} and \ref{tab:main-table2} present the results for the present and absent keyphrase performances, respectively, for our different experiments. The results demonstrate that adding extra information from the body of the given article to the baseline (T+A) is beneficial for both present and absent keyphrase generation. Augmenting the summary sentences provides the most substantial boost compared to any other method for both present and absent keyphrase performance. For instance, improvements in present keyphrase performance on F1@M metric are: catSeq (0.338 $\rightarrow$ 0.380), ExHiRD (0.325 $\rightarrow$ 0.366), LED (0.360 $\rightarrow 0.397$). The absent keyphrase performance using the summary method improves up to four times the baseline performance; for example, catSeq (0.021 $\rightarrow$ 0.079), ExHiRD (0.030 $\rightarrow$ 0.080), LED (0.061 $\rightarrow$ 0.106). Interestingly, with One2Set, although F1@M for present keyphrase generation does not improve with Summary and Random methods, the trend for absent keyphrase performance is very similar to that of the other models. The discrepancy in the present keyphrase generation performance could be because One2Set might potentially be biased towards identifying the keyphrases from the earlier portions of the documents. For our setting of long documents, it may require better heuristics to initialize the control codes so as to attend to the later portions of the document.  
Still, in three out of four models, adding summary information gives us the best results. This seems to follow the natural intuition that the summary of the article contains the most topical information useful for the generation the keyphrases. 

\begin{figure}[t]
    \centering
    \includegraphics[width=1.0\linewidth, height=5.0cm]{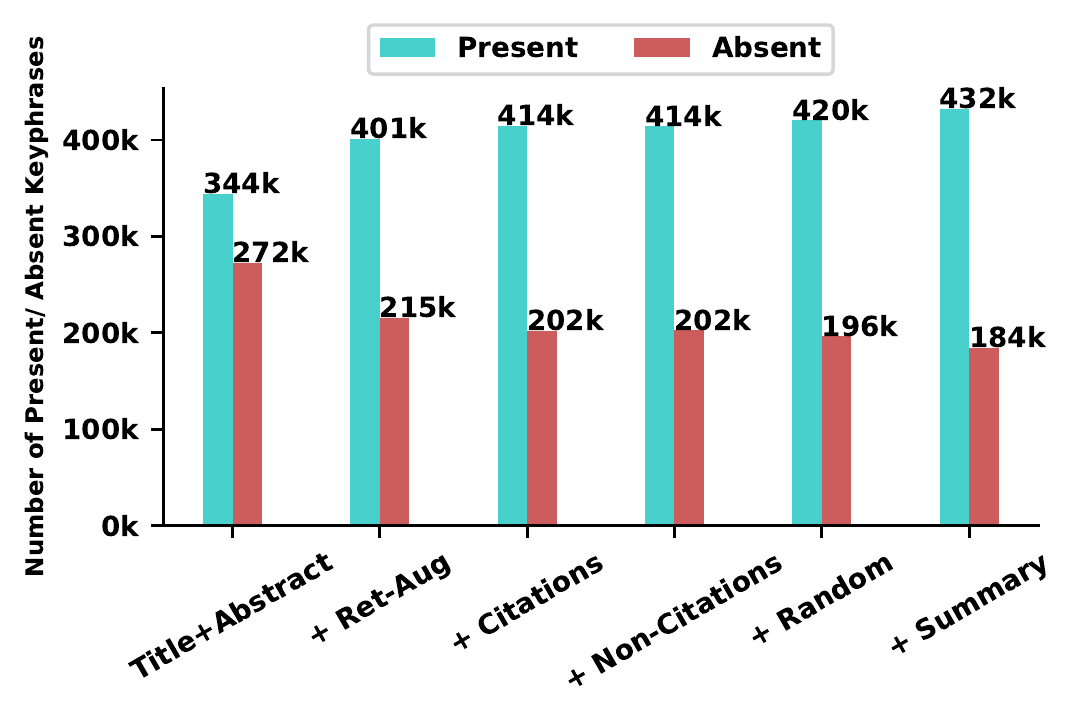}
    \caption{\#Present and \#Absent Keyphrases in terms of the whole input text (for various methods) in \textsc{FulltextKP} dataset. $+$ indicates concatenation of title and abstract with the specified information.}
    \label{fig:plot_stats}
\end{figure}

Figure \ref{fig:plot_stats} further validates the above intuition. Summary method is the richest source of present keyphrases. The total number of present keyphrases in the texts of all Summary samples is about $\sim$88,000 higher than just Title+Abstract samples. So, the model learns to generalize better when trained on such highly topical input texts of Summary method.

Interestingly, we find that using only the citation sentences results in worse performance compared to using non-citation sentences or just random sentences. This observation is in contrast to the observation of the use of citation sentences for the keyphrase extraction methods. We hypothesize the reason for such an observation is that keyphrase extraction methods focus mainly on identifying keyphrases by capturing the statistical information such as tf-idf or word co-occurrences that appear in the target document and the cited ones (through the citation sentences), whereas the deep learning methods try to get a semantic or holistic understanding of the document itself (i.e., from all sentences).

Incorporating either non-citation sentences or random sentences gives very similar results because the entire text contains majority of non-citation sentences. Unfortunately, appending the sentences retrieved from the semantically similar documents (using \textsc{SPECTER}) performs worse than the baseline. We hypothesize that even though the sentences have semantic similarity on the embedding level, they could still be significantly different in terms of their meaning and may not contain exactly the same keyphrases. Rather, a lot of such sentences may confuse and deviate the model from predicting the gold keyphrases.

Among the four different models, Longformer Encoder-Decoder (LED) performs the best. We hypothesize that this happens because it is based on the more sophisticated Transformer model architecture and also because it is suited for reading long documents. However, LED training takes at least 5-10 times more than the catSeq model. We provide more details about the compute power and time in Appendix \ref{sec:app-time-and-memory}.

We performed a 2-tailed statistical significance test with alpha\footnote{Probability of rejecting the null hypothesis when it was actually true} value as 0.05 and the p-values\footnote{Probability of getting a result that is as extreme or more extreme when the null hypothesis is true} less than 0.05 were considered statistically different. We observed that the results were statistically different for all models except One2Set. With One2Set, the additional information to the T+A baseline does not yield statistically different results.

\begin{table}[t]
\centering
\small
\renewcommand{\arraystretch}{1.3}
\setlength\tabcolsep{2.6pt}
\begin{tabular}{lllllll}
\hline
\Xhline{2\arrayrulewidth}
Method      & P@5 & R@5 & F1@5 & P@M & R@M & F1@M  \\
\hline
Abs\_Sum & 0.201$_1$ & 0.405$_2$ & 0.268$_1$ & 0.326$_4$ & 0.409$_2$ & 0.363$_3$ \\
Ext\_Sum & \textbf{0.217$_4$} & \textbf{0.431$_8$} & \textbf{0.289$_5$} & \textbf{0.365$_2$} & \textbf{0.435$_8$} & \textbf{0.397$_3$} \\
\hline
Abs\_Sum & 0.020$_1$ & 0.048$_2$ & 0.029$_1$ & 0.062$_1$ & 0.048$_2$ & 0.054$_1$ \\
Ext\_Sum & \textbf{0.047$_1$} & \textbf{0.100$_1$} & \textbf{0.064$_1$} & \textbf{0.112$_4$} & \textbf{0.100$_1$} & \textbf{0.106$_2$} \\

\hline
\Xhline{2\arrayrulewidth}
\end{tabular}
\caption{\label{tab:exp_cls} Results comparing the performance of Abstractive Summarization vs. Extractive Summarization methods. Subscripts denote the standard deviation as multiples of $\pm$ 0.001.}
\label{tab:abs_vs_ext}
\end{table}

\begin{table*}[ht]
\small
\centering
\renewcommand{\arraystretch}{1.3}
\setlength\tabcolsep{11pt}
\newcommand\mc[1]{\multicolumn{1}{l}{#1}}
\begin{tabular}{c|cccccccc}
\hline
\Xhline{3\arrayrulewidth}
 \mc{Type} & Method & Seq-Len & P@5 & R@5 & F@5 & P@M & R@M & F1@M \\
 \hline
\multirow{5}{*}{\makecell{ Present \\ Keyphrases}} & Summary & 800 & \textbf{0.217$_4$} & \textbf{0.431$_8$} & \textbf{0.289$_5$} & 0.365$_2$ & \textbf{0.435$_8$} & 0.397$_3$ \\
\cline{2-9}
& \multirow{4}{*}{Random} & 800 & 0.213$_0$ & 0.428$_1$ & 0.284$_0$ & 0.354$_6$ & 0.431$_0$ & 0.389$_3$\\
 &  & 1500 & 0.211$_2$ & 0.424$_3$ & 0.282$_2$ & 0.370$_3$ & 0.427$_3$ & 0.397$_1$ \\
 &  & 2000 & 0.215$_2$ & 0.428$_3$ & 0.286$_2$ & 0.372$_4$ & 0.431$_3$ & 0.399$_4$ \\
 &  & 2500 & 0.212$_4$ & 0.425$_6$ & 0.283$_5$ & \textbf{0.389$_1$} & 0.427$_7$ & \textbf{0.407$_3$} \\
\hline
\multirow{5}{*}{\makecell{Absent \\ Keyphrases}} & Summary & 800 & 0.047$_1$ & 0.100$_1$ & 0.064$_1$ & 0.112$_4$ & 0.100$_1$ & 0.106$_2$ \\
\cline{2-9}
& \multirow{4}{*}{Random} & 800 & 0.036$_2$ & 0.078$_4$ & 0.049$_2$ & 0.093$_1$ & 0.078$_4$ & 0.085$_2$ \\
 &  & 1500 & 0.044$_0$ & 0.097$_1$ & 0.061$_1$ & 0.110$_2$ & 0.097$_1$ & 0.103$_1$ \\
 &  & 2000 & 0.047$_2$ & 0.100$_3$ & 0.063$_2$ & 0.114$_2$ & 0.100$_3$ & 0.107$_3$ \\
 &  & 2500 & \textbf{0.049$_2$} & \textbf{0.108$_4$} & \textbf{0.067$_2$} & \textbf{0.126$_2$} & \textbf{0.108$_4$} & \textbf{0.116$_3$} \\ 
 \hline
 \Xhline{3\arrayrulewidth}
\end{tabular}
\caption{Results comparing the performance of Longformer Encoder-Decoder Model with source texts of different sequence lengths. Subscripts denote the standard deviation as multiples of $\pm$ 0.001.}
\label{tab:seq-len}
\end{table*}

\subsection{Analysis on Summarization Methods}
\label{sec:summarization}
In this section, we further investigate the capability of our best performing method, i.e., extractive summarization, for keyphrase generation in comparison with an abstractive summarization approach, abbreviated as \textit{T+A+Abs\_Sum} or simply \textit{Abs\_Sum}.
%
For the abstractive summarization,  we use BigBirdPegasus \cite{zaheer2020big} model, pretrained on ArXiv dataset \cite{cohan2018discourse} to first generate the abstractive summaries. Next, we augment the sentences from abstractive summaries to the title and abstract of the articles. Further, we train our best performing model (i.e., LED) using \textit{T+A+Abs\_Sum} method. In Table \ref{tab:abs_vs_ext}, we compare the results of \textit{T+A+Abs\_Sum} with the \textit{T+A+Summary} (extractive) method. Note that, by default, we use Summary for the Extractive Summarization method \cite{zheng2019sentence}, also abbreviated as \textit{Ext\_Sum}. 

From Table \ref{tab:abs_vs_ext}, we observe that the extractive summary added to T+A performs better than the abstractive summary (added to T+A) by 0.034 (F1@M) and 0.052 (F1@M) for present and absent keyphrase generation, respectively. 
The behavior could be explained as follows: (1) Extractive summaries contain sentences that are more central to the paper \cite{zheng2019sentence}. Thus, they tend to bring complementary information from all parts of the paper that improves the performance of keyphrase generation models; and (2) We manually inspected a small subset of the dataset and observed that about two-thirds of the sentences in the abstractive summaries were from the Introduction section of the papers, which although contain extended but potentially overlapping or paraphrased description of the Abstract section. Further, we observed that there were about 10-15\% repetitive statements in the abstractive summaries. We conclude that augmenting repetitive statements or statements with redundant information as the abstract do not aid the keyphrase generation performance. 

\begin{table*}[ht!]
\small
\centering
\def\columnseprulecolor{\color{grey}}
\resizebox{\textwidth}{!}{
  \begin{tabular}{p{42em} | p{11.5em}}
    \toprule
   \multicolumn{1}{c}{\rule{0pt}{2ex}\textbf{Snippets from title and abstracts and summary with model predictions}} &   \multicolumn{1}{c}{\textbf{Gold Keyphrases}}\\[1ex] 
    \hline
    \addlinespace[0.1cm]
    \rule{0pt}{3ex}\colorbox{cyan!25!white}{Segmentation} of \colorbox{cyan!25!white}{brain MR images} using \colorbox{cyan!25!white}{intuitionistic fuzzy clustering} algorithm $<$sep$>$ A new algorithm ... using \colorbox{cyan!25!white}{intuitionistic fuzzy clustering} (\colorbox{cyan!25!white}{IFCM}), is proposed in this paper. & segmentation, brain mr image, intuitionistic fuzzy set\\[0.5ex]
    \hline
     \rule{0pt}{3ex}\colorbox{cyan!25!white}{GPU}-based simulation of  \colorbox{cyan!25!white}{wireless body area network} $<$sep$>$ ... in a local workstation to perform  \colorbox{cyan!25!white}{model simulations} of a ... the goal of this project is to gain an understanding of the \colorbox{cyan!25!white}{gpu}-based \colorbox{cyan!25!white}{performance} of wban \colorbox{cyan!25!white}{model simulation} ... &  gpu, wireless body area network, model, matlab\\[0.5ex]
    \hline
    \rule{0pt}{3ex}Spectrum sharing for \colorbox{cyan!25!white}{directional systems} $<$sep$>$ \colorbox{cyan!25!white}{Dynamic spectrum access systems} will need to ... An example involving a satellite downlink antenna and a broadband wireless access system using \colorbox{cyan!25!white}{directional antennas} is presented ... & dynamic spectrum access, directionality\\[0.5ex]
    \hline
    \rule{0pt}{3ex}Automatic classification of \colorbox{cyan!25!white}{anuran} sounds using \colorbox{cyan!25!white}{convolutional neural networks} $<$sep$>$ ... networks with mel-frequency cepstral coefficients (\colorbox{cyan!25!white}{mfccs}) as input for the task of ... $<$sep$>$ Different \colorbox{cyan!25!white}{machine learning} approaches for \colorbox{cyan!25!white}{anuran} classification ... & anuran, convolutional neural networks, wireless sensor networks, mfcc, machine learning\\
    \bottomrule
    \addlinespace[0.1cm]
  \end{tabular}
}

  \caption{Predicted keyphrases by Longformer Encoder-Decoder model are highlighted in \colorbox{cyan!25!white}{cyan}. The source text is in the format: Title $<$sep$>$ Abstract $<$sep$>$ Summary-Sent\_1 ... $<$sep$>$ Summary-Sent\_k}
    \label{tab:predictions}
\end{table*}

\vspace{2mm}
\subsection{Exploring longer sequence lengths with Longformer Encoder-Decoder}
\vspace{2mm}

\label{sec:LED-exploration}
Scientific documents are often very long sequences of text, ranging from a few thousands to tens of thousands of tokens. Longformer Encoder-Decoder, proposed by \citet{Beltagy2020Longformer}, has the capability to deal with long sequences of text. We therefore explore the effect of sequence lengths longer than 800 on the present and absent keyphrase performance. We prepare different versions of the \textit{T+A+Random} with the maximum sequence lengths of 800, 1500, 2000 and 2500. We chose the \textit{Random} version instead of directly using the entire source document since \textit{Random}  would be a better representative of the long document if we need to truncate the sequence length to a fixed budget (i.e., rather than truncating and completely removing a part of the document, we ensure coverage of sentences from the entire document). Note that these sequence length values are based on the word-level tokenization based on NLTK's word-tokenizer. The sequence length value based on subword tokenization (used by Transformers) is actually much longer, but fortunately, LED is able to handle sequence length up to 16000 tokens.

The results in Table \ref{tab:seq-len} show that both present and absent keyphrase performance generally increases as we increase the source text length. In the table, we also compare the performance of these different versions of \textit{T+A+Random} with the \textit{T+A+Summary} with a much smaller sequence length of 800, so as to justify the purpose of smaller intended summary. We observe that the performance of \textit{Summary-800} is quite comparable to \textit{Random-2000} for both present and absent keyphrase performance, particularly for the most important metric F1@M. This further validates our hypothesis that summary indeed gives us a richer source of information. Just the sequence length of 800 can condense the information that could otherwise be extracted using sequence length of 2000. Furthermore, the summary has the benefit of fitting a smaller memory and computational resources budget. We observed that with the increasing sequence lengths, the training time increased at least at a linear rate and used considerable amount of GPU and RAM memory. In the case we add more sentences to the summary, we expect the quality to degrade and converge to that of the \textit{Random} versions and furthermore, it will no more stand out as a summary of the document. 

In Table \ref{tab:predictions}, we show sample predictions for the best performing method \textit{T+A+ (Extractive) Summary} using LED model. The gold keyphrases and the predicted keyphrases are tokenized and stemmed before comparison. We make the following observations from the table. First, the model can predict accurately both the present and absent keyphrases. Second, some predicted keyphrases are near-matches of the gold keyphrases, and despite their quality as keyphrases, due to the limitation of the metric (exact match), they do not contribute to the model performance.

\section{Conclusion}
\vspace{-2mm}
In this paper, we explore numerous ways of incorporating additional data from the body of the scholarly documents for improving the performance of the keyphrase generation task. We conclude from the results that the extractive summary sentences, that are more central to the paper, provide the most topical information for boosting both present and absent keyphrase generation performance. The citation, non-citation, and random sentences also bring complementary information to improve the performance modestly. Some other ways such as augmenting semantically similar sentences from other papers or augmenting abstractive summary sentences, that bring repetitive or redundant information to titles and abstracts, are not  effective. We present a comprehensive analysis with four models including LED, yet unexplored for the task. Our work aims at breaking apart the barriers of using only titles and abstracts, and presents a large-scale dataset with full texts for keyphrase generation. 


\section{Limitations}
One of the limitations of the proposed methods is the increased (up to 2-3 times) compute time and memory for the training of the models compared to the conventional training of the models using only T+A of the articles. We provide more details in Appendix \ref{sec:app-time-and-memory}. Further, our best performing method \textit{T+A+Summary} requires an additional modest overhead of pre-computing summaries for all the articles.

Another potential limitation of our work is that we can not directly compare our models on the widely used datasets, e.g., KP20k, Inspec, Krapivin, NUS since these datasets do not have full texts of the papers. To be comprehensive, we considered the performance of four models on the new dataset.

We encourage future work in the direction of better ways of integrating external information including more sophisticated approaches for the summarization of scientific documents. 

\section{Ethical Considerations}
Since keyphrase generation finds direct application in many downstream tasks such as recommender systems, reviewer matching, articles clustering for fast retrieval, etc., it is important to consider the ethical implications of these models. The models are not perfect and might make misleading predictions at times. The users must make their own discrimination in deploying the current models for decision-making systems. 


\section*{Acknowledgements} 
This research is supported in part by NSF CAREER
award \#1802358, NSF CRI award \#1823292, and UIC Discovery Partners Institute (DPI) award.
Any opinions, findings, and conclusions expressed
here are those of the authors and do not necessarily
reflect the views of NSF or DPI. We thank AWS for computational resources used in this study. We also thank our anonymous
reviewers for their constructive feedback and suggestions.

\bibliography{acl}
\bibliographystyle{acl_natbib}

\appendix
\clearpage
\section{Hyperparamters}
\label{sec:hyperparameters}

For the catSeq model, we use the same hyperparameters as \citet{chan-etal-2019-neural}. For ExHiRD and One2Set, we use the respective repositories with their respective hyperparameters. For the LED model, we did a hyperparameter sweep over the different learning rates $\in$ \{1e-3, 1e-4, 1e-5, 2e-5, 5e-5, 8e-5\} and finally chose the learning rate as 5e-5, optimizer as Adam \cite{kingma2014adam}, train batch size as 12. For preprocessing the articles to generate summary using PACSUM (TF-IDF), we use the default hyperparameters $\beta = 0$, $\lambda_1 = 0$, $\lambda_2 = 1$. We ran each experiment thrice with different random seeds using V100 (16GB) and A6000 (48GB) GPU machines.

\section{Time and Memory Consumption}
\label{sec:app-time-and-memory}
In Table \ref{tab:time_mem}, we provide the comparison of time and memory utilization between the conventional (i.e., Title+Abstract) and new methods (i.e., Title+Abstract+Body) in reference to our best performing model (i.e., LED). Body refers to sentences retrieved using any of the methods described in \S\ref{sec:methods}, i.e., Random, Citation, Non-Citation, Summary, Retrieval-Augmentation.

The conventional method takes about 3.6 hours/ epoch and 10 GB of CUDA memory with A6000 GPU for training. Whereas the newer methods take 5.5 hours/ epoch and about 30 GB of CUDA memory for training. The inference time is less than 1 GPU hour for both types of methods.

\section{PacSum Algorithm}
\label{sec:PacSum}


\begin{table}
\small
\renewcommand{\arraystretch}{1.3}
\setlength\tabcolsep{10pt}
\centering
\begin{tabular}{ccc}
\hline
\Xhline{2\arrayrulewidth}
Methods & Time & CUDA Memory \\
\hline
Title+Abstract    & 3.6 hrs/ epoch & 10 GB \\
+Body & 5.5 hrs/ epoch & 30 GB\\
\hline
\Xhline{2\arrayrulewidth}
\end{tabular}
\caption{\label{tab:exp_cls} Time and Memory consumption using the conventional method (T+A) and the new methods (T+A+Body) where Body could be sentences retrieved using any of the methods like Summary, Citation, Non-Citation, Random, Ret-Aug, with total sequence length 800.}
\label{tab:time_mem}
\vspace{+2em}
\end{table}

\begin{algorithm}[t]
\begin{small}
\caption{Summarization: PacSum (TF-IDF)}
\textbf{Input} $\Phi$: Document split into a list of sentences $(s_i)_{i \in [1, N]}$
(tokenized into list of words); hyperparameters: $\beta$; $\lambda_1$; $\lambda_2$\\
\textbf{Output} $\mathbb{S}$: Summary of document $\Phi$
\vspace{0.2em}
\begin{algorithmic}[1]
\State $N \gets \text{Number of sentences in document}$
\vspace{0.2em}
\State $\text{IDF} \gets \Big[\log(N - \text{count}(w, \Phi) + 0.5) - \log(\text{count}(w, \Phi) + 0.5) \textbar w \in s_i; s_i \in \Phi \Big]$
\vspace{0.2em}
\State $\text{TF} \gets \Big[\big[\text{count}(w, s_i) \textbar w \in s_i \big] \textbar s_i \in \Phi\Big]$
\vspace{0.2em}
\State $\forall i, j \in [1, N], \text{sim}(i,j) \gets \newline \left\{
    \begin{array}{cl}
        \mathbf{1} & \mbox{i == j} \\
        \underset{w \in s_i \& s_j}{\sum} \Big(\text{TF}[s_i][w] * \text{TF}[s_j][w] * \text{IDF}[w] \Big) ^2 & \mbox{i != j}
    \end{array}
\right.$
\vspace{0.2em}
\State $\tau = \min(\text{sim}) + \beta * (\max(\text{sim}) - \text{min}(\text{sim}))$
\vspace{0.2em}
\State $\forall i, j \in [1, N], \text{sim}(i, j) = \text{sim}(i, j) - \tau$
\vspace{0.2em}
\State $\underset{\forall i \in [1, N]}{\text{FS}(i)} \gets \left\{
    \begin{array}{cl}
        \underset{\forall j, j < i}{- \sum} \text{sim}(i, j) & \text{sim}(i, j) > \tau\\
        0 & \text{otherwise}
    \end{array}
\right.$
\vspace{0.2em}
\State $\underset{\forall i \in [1, N]}{\text{BS}(i)} \gets \left\{
    \begin{array}{cl}
        \underset{\forall j, j > i}{\sum} \text{sim}(i, j) & \text{sim}(i, j) > \tau\\
        0 & \text{otherwise}
    \end{array}
\right.$
\vspace{0.2em}
\State $\text{centrality}(i) \gets \lambda_1 * \text{FS}(i) + \lambda_2 * \text{BS}(i), \forall i \in [1, N]$
\vspace{0.2em}
\State $\text{indices} \gets \text{argsort}(\text{centrality, desc=True})$
\vspace{0.2em}
\State $\mathbb{S} \gets \Phi[\text{indices}]$ 
\vspace{0.2em}
\State $\text{Return } \mathbb{S}$

\end{algorithmic}
\label{alg:summmary}
\end{small}
\end{algorithm}
\vspace{2mm}

\section{Retrieval-Augmentation Algorithm}
\label{sec:ret-aug}
\begin{algorithm}[t]
\setlength{\intextsep}{1\baselineskip}
\begin{small}
\caption{Retrieval-Augmentation} 
\textbf{Input} Collection of scientific articles $\mathbb{C}$\\
\textbf{Output} Collection of scientific articles $\mathbb{D}$ with Title $\oplus$ Abstract $\oplus$ Semantically similar sentences
\begin{algorithmic}[1]
\State $\Phi \gets \underset{a \in \mathbb{C}_{\text{train}}}{\bigcup}\{\underset{s_j \in (T_a \oplus A_a)}{\bigcup}s_j\}$ \Comment{$T_a$: Title, $A_a$: Abstract, $\Phi$: Set of sentences from titles and abstracts of all articles in training dataset}
\vspace{0.2em}
\State $\mathbb{D} \gets \varnothing$
\ForAll {$a \in \mathbb{C} $}
\vspace{0.2em}
\State $\Phi_q \gets  \underset{s_j \in (T_a \oplus A_a)}{\bigcup} s_j$
\vspace{0.2em}
\State $\Phi_{\text{key}} \gets \Phi - \Phi_q$
\vspace{0.2em}
\State $E_{\text{key}} \gets \big(\text{embed}(s_i)\big)_{ s_i \in \Phi_{\text{key}}}$
\vspace{0.2em}
\State $\vec e_q \gets \text{embed}(T_a \oplus A_a)$
\vspace{0.2em}
\State $ \text{similarity} \gets \big(\vec e_{\text{key}} \cdot \vec e_q\big )_{\vec e_{\text{key}} \in E_{\text{key}}}$
\vspace{0.2em}
\State $\text{indices} \gets \mathrm{argmax_k}(\text{similarity})$
\vspace{0.2em}
\State $\text{extra\_info} \gets \big(\Phi_{\text{key}}[i]\big )_{i \in \text{indices}}$
\vspace{0.2em}
\State $a \gets T_a \oplus A_a \oplus \text{extra\_info}$
\vspace{0.2em}
\State $\mathbb{D} \gets \mathbb{D} \cup \{a\}$
\vspace{0.2em}
\EndFor
\vspace{0.2em}
\State $\text{Return  } \mathbb{D}$
\end{algorithmic}
\label{alg:ret-aug}
\end{small}
\end{algorithm}

\end{document}